\documentclass[conference]{IEEEtran}
\IEEEoverridecommandlockouts
\usepackage{cite}
\usepackage{amsmath,amssymb,amsfonts}
\usepackage{algorithmic}
\usepackage[ruled,vlined,linesnumbered]{algorithm2e}
\usepackage{tabularray}
\usepackage{multicol}
\usepackage{multirow}
\usepackage{tabu}
\usepackage{diagbox}
\usepackage{float}
\usepackage{booktabs}
\usepackage{makecell}
\usepackage{subfig}
\usepackage{graphicx}
\usepackage{caption}
\usepackage{textcomp}
\usepackage{hyperref}
\usepackage{url}
\usepackage{xcolor}
\def\BibTeX{{\rm B\kern-.05em{\sc i\kern-.025em b}\kern-.08em
    T\kern-.1667em\lower.7ex\hbox{E}\kern-.125emX}}
\usepackage{bm}

\makeatletter
\renewcommand{\maketag@@@}[1]{\hbox{\m@th\normalsize\normalfont#1}}%
\makeatother
\begin{document}

\title{Accelerating Scalable Graph Neural Network Inference with Node-Adaptive Propagation\\
}
\author{
    \IEEEauthorblockN{Xinyi Gao$^{1}$, Wentao Zhang$^{2,3*}$, Junliang Yu$^1$, Yingxia Shao$^{4}$, Quoc Viet Hung Nguyen$^{5}$, Bin Cui$^{2}$, Hongzhi Yin$^{1*}$}
    \IEEEauthorblockA{$^1$ The University of Queensland, Brisbane, Australia}
    \IEEEauthorblockA{$^2$ Peking University, Beijing, China}
     \IEEEauthorblockA{$^3$ National Engineering Laboratory for Big Data Analysis and Applications, Beijing, China}
    
    \IEEEauthorblockA{$^4$ Beijing University of Posts and Telecommunications, Beijing, China}
    \IEEEauthorblockA{$^5$ Griffith University, Gold Coast, Australia}

\thanks{* Corresponding authors. E-mail addresses: h.yin1@uq.edu.au (Hongzhi Yin) and wentao.zhang@pku.edu.cn (Wentao Zhang)}
}

\maketitle

\begin{abstract}

Graph neural networks (GNNs) have exhibited exceptional efficacy in a diverse array of applications. However, the sheer size of large-scale graphs presents a significant challenge to real-time inference with GNNs. Although existing Scalable GNNs leverage linear propagation to preprocess the features and accelerate the training and inference procedure, these methods still suffer from scalability issues when making inferences on unseen nodes, as the feature preprocessing requires the graph to be known and fixed. To further accelerate Scalable GNNs inference in this inductive setting, we propose an online propagation framework and two novel node-adaptive propagation methods that can customize the optimal propagation depth for each node based on its topological information and thereby avoid redundant feature propagation. The trade-off between accuracy and latency can be flexibly managed through simple hyper-parameters to accommodate various latency constraints. Moreover, to compensate for the inference accuracy loss caused by the potential early termination of propagation, we further propose Inception Distillation to exploit the multi-scale receptive field information within graphs. The rigorous and comprehensive experimental study on public datasets with varying scales and characteristics demonstrates that the proposed inference acceleration framework outperforms existing state-of-the-art graph inference acceleration methods in terms of accuracy and efficiency. Particularly, the superiority of our approach is notable on datasets with larger scales, yielding a $75\times$ inference speedup on the largest Ogbn-products dataset.

\end{abstract}


\section{Introduction}

Developing a graph neural network (GNN) for very large graphs has drawn increasing attention due to the powerful expressiveness of GNNs and their enormous success in many industrial applications~\cite{hu2021ogb, yi2022flag,chen2021uniting, yu2023self}. 
While GNNs provide a universal framework to tackle various down-streaming tasks, their implementation on large-scale industrial graphs is impeded by heavy computational demands, which severely limits their application in latency-sensitive scenarios.
For example, recommender systems designed for streaming sessions must completely perform real-time inference on user-item interaction graphs \cite{chandramouli2011streamrec, zheng2023personalized, zheng2022automl, wu2019session, li2021lightweight, zhang2021graph, xia2023efficient,duong2021efficient, chen2023neural, xia2023towards}. The fraud and spam detection tasks require millisecond-level inference on the million-scale graph to identify the malicious users and prevent the loss of assets for legitimate users \cite{wang2019semi, 101145, liu2018heterogeneous}. In some computer vision applications, GNNs are designed for processing 3D point cloud data and deployed in the automated driving system to perform object detection or semantic segmentation tasks \cite{qi20173d, shi2020point, landrieu2018large}. In such scenarios, the real-time inference is of utmost importance.

\begin{figure}[ht]
\includegraphics[width=\linewidth]{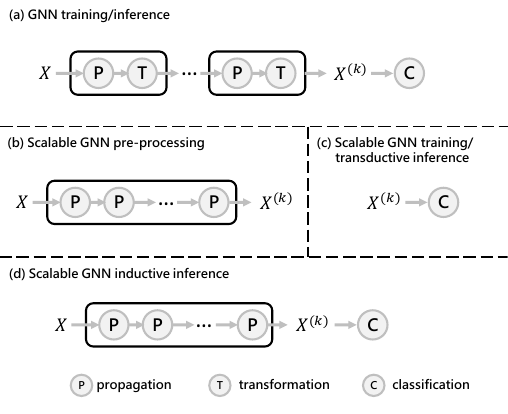}
\caption{The comparison of GNN and Scalable GNN in the training and inference procedures. (a) The training and transductive/inductive inference procedure of GNNs, such as GCN \cite{DBLP:conf/iclr/KipfW17}. The propagation and transformation process are integrated as a single GNN layer. (b) The feature preprocessing approach of Scalable GNNs, illustrated by SGC \cite{wu2019simplifying}. The transformation processes are removed, and non-parametric propagation processes are conducted as the preprocessing. (c) The training/transductive inference procedure of Scalable GNNs. The classifier training or transductive inference is based on preprocessed features directly. (d) The inductive inference procedure of Scalable GNNs, where propagation has to be executed online.}
\label{fig_pre}
\end{figure}

The root cause for the heavy computation and high latency in GNNs is known as the \textit{neighbor explosion problem}.
GNNs~\cite{DBLP:conf/iclr/KipfW17, hamilton2017inductive, gao2023semantic, DBLP:conf/iclr/VelickovicCCRLB18} typically adopt the message-passing pipeline and leverage the feature propagation and transformation processes to construct the model as shown in Figure \ref{fig_pre} (a). Through executing $k$ feature propagation processes, the propagated features at depth $k$ can capture the node information from $k$-hop neighborhoods (also known as supporting nodes). This approach is of particular significance to large-scale and sparsely labeled graphs, where large propagation depths are needed to aggregate enough label information from distant neighbors \cite{hu2020ogb, zeng2021decoupling}. However, as the depth of propagation increases, the number of supporting nodes grows exponentially, incurring a substantial computational cost.

To mitigate the expensive computation resulting from feature propagation, several Scalable GNNs~\cite{wu2019simplifying, sign_icml_grl2020, zhang2022pasca, chen2020scalable, zhang2022nafs, zhu2020simple, zhang2022graph}, e.g., SGC, were proposed to remove the non-linear transformation layers among propagation layers and use the propagated feature for prediction. 
As shown in Figure \ref{fig_pre} (b), the feature propagation of these methods can be pre-computed in the preprocessing procedure and only needs to be executed once. Instead of performing feature propagation during each training epoch, the training time complexity of Scalable GNNs is significantly reduced than other GNNs \cite{bojchevski2020pprgo, gasteiger2018predict}, and the training of these models scales well with graph size (Figure \ref{fig_pre} (c)). Due to their notable advantage in training speed, Scalable GNNs have gained considerable popularity in large-scale graph processing tasks. This is evident from the widespread adoption of Scalable GNNs as the primary choice in numerous methods participating in the Open Graph Benchmark challenge \cite{ogb}. 

However, Scalable GNNs still struggle to do efficient inference on unseen nodes because the preprocessing of feature propagation is based on the transductive premise (i.e. the graph is known and fixed). 
In the majority of practical situations, inference encompasses unseen nodes, requiring online execution of feature propagation~\cite{hamilton2017inductive, DBLP:conf/iclr/ZengZSKP20}, as depicted in Figure \ref{fig_pre} (d). This condition significantly hinders real-world applications.
In addition, existing methods adopt a fixed propagation depth for all nodes, which restricts the flexibility of exploiting the multi-scale features and tends to over-smooth the high-degree nodes \cite{zhang2021node}, thereby leading to redundant computation and performance degradation. 

In this paper, we aim to reduce the redundant computation of feature propagation to further accelerate the inference of Scalable GNNs in the inductive setting. 
To this end, we propose a general online propagation framework: Node-Adaptive Inference (NAI), which introduces \textit{personalized propagation depth} to determine the optimal propagation depth for inference. Specifically, we provide two different node-adaptive propagation approaches within the NAI framework, which could evaluate the smoothing status of the propagated feature in an explicit and implicit manner, respectively, and thus can terminate the needless propagation in time. Particularly, the trade-off between inference latency and accuracy can be flexibly managed by tuning simple global hyper-parameters, which adaptively adjust the propagation depth for each node. This provides a variety of inference options for users with different latency constraints. Moreover, taking inspiration from the Inception Network \cite{szegedy2015going}, which is the cornerstone work in convolutional neural networks, we design a novel Inception Distillation method in NAI to exploit the multi-scale receptive field information and mitigate the performance degradation caused by the adaptive propagation. With a more powerful supervision signal, NAI could accelerate the inference speed with a negligible performance drop.

The main contributions of this paper are summarized as follows:
\begin{itemize}
  \item \textbf{New challenge.} We focus on the inference acceleration in a more real and challenging setting - graph-based inductive inference, where the ever-Scalable GNNs also struggle with heavy online computation of feature propagation.
  \item \textbf{New Methodology.} We propose an online propagation framework for Scalable GNNs and two novel node-adaptive propagation approaches that generate the personalized propagation depth for each node, thereby avoiding the redundant computation of feature propagation and mitigating the over-smoothing problem. To compensate for the potential inference accuracy loss, we further propose Inception Distillation to exploit the multi-scale receptive field information and improve the accuracy.
  \item \textbf{SOTA Performance.}   Extensive experiments are conducted on three public datasets with different scales and characteristics, and the experimental results show that our proposed efficient inference framework NAI outperforms the state-of-the-art (SOTA) graph inference acceleration baselines in terms of both accuracy and efficiency. In particular, the advantage of our NAI is more significant on larger-scale datasets, and NAI achieves $75\times$ inference speedup on the largest Ogbn-products dataset.
\end{itemize}

\section{Preliminary}
\subsection{Problem Formulation}
Given a graph $\mathcal{G}$ = ($\mathcal{V}$, $\mathcal{E}$) with $|\mathcal{V}| = n$ nodes and $|\mathcal{E}| = m$ edges, its node adjacency matrix and degree matrix are denoted as ${\mathbf{A}} \in \mathbb{R}^{n \times n}$ and $\mathbf{D} = \mathrm{diag}(d_1, d_2, ..., d_n$), where $d_i =  {\textstyle \sum_{v_j\in \mathcal{V}}\mathbf{A}_{i,j}} $ is the degree of node $v_i \in \mathcal{V}$. The adjacency matrix and degree matrix with self-loops are denoted as $\widetilde{\mathbf{A}} $ and $\widetilde{\mathbf{D}}$. The node feature matrix is $\mathbf{X} = \{\boldsymbol{x}_1, \boldsymbol{x}_2, ..., \boldsymbol{x}_n\}$ in which $\boldsymbol{x}_i\in\mathbb{R}^{f}$ represents the node attribute vector of $v_{i}$, and $\mathbf{Y} = \{\boldsymbol{y}_1, \boldsymbol{y}_2, ..., \boldsymbol{y}_l\}$ is the one-hot label matrix for node classification task. 
In the inductive setting, the entire node set $\mathcal{V}$  is partitioned into training set $\mathcal{V}_{train}$ (including labeled set $\mathcal{V}_l$ and unlabeled set $\mathcal{V}_u$) and test set $\mathcal{V}_{test}$. GNNs are trained on $\mathcal{G}_{train}$ which only includes $\mathcal{V}_{train}$ and all edges connected to $v \in \mathcal{V}_{train}$. The evaluation is to test the performance of trained GNNs on $\mathcal{V}_{test}$ in graph $\mathcal{G}$.

\subsection{Graph Neural Networks}
GNNs aim to learn node representation by using topological information and node attributes. The existing GNNs adopt the message-passing pipeline and construct models utilizing two processes: feature propagation and transformation. By stacking multiple layers, the $k$-th layer feature matrix $\mathbf{X}^{(k)}$ can be formulated as:
\begin{equation}
\begin{aligned}
&\mathbf{X}^{(k)} =\delta\left(\hat{\mathbf{A}}\mathbf{X}^{(k-1)}\mathbf{W}^{(k)}\right), \\
&\hat{\mathbf{A}}=\widetilde{\mathbf{D}}^{\gamma-1}\widetilde{\mathbf{A}}\widetilde{\mathbf{D}}^{-\gamma},   
\end{aligned}
\label{eq_GCN}
\end{equation}
where $\mathbf{W}^{(k)}$ is the layer-specific trainable weights at layer $k$ and $\delta\left(\cdot\right)$ is the activation function, such as ReLU function. $\widetilde{\mathbf{D}}$ is the diagonal node degree matrix used to normalize $\widetilde{\mathbf{A}}$.
$\mathbf{X}^{(k-1)}$ is the input of $k$-th layer and $\mathbf{X}^{(0)}=\mathbf{X}$.
In each layer, $\hat{\mathbf{A}}$ propagates the information among neighbors, and $\mathbf{W}^{(k)}$ transforms the propagated features. 
Through executing $k$ times feature propagation processes, the propagated features at depth $k$ can capture the node information from $k$-hop neighborhoods. However, as the depth of propagation increases, the number of supporting nodes grows exponentially, incurring a substantial computational cost.
Note that, $\gamma \in [0, 1]$ is the convolution coefficient and could generalize Eq. (\ref{eq_GCN}) to various existing models. By setting $\gamma=1$, 0.5 and 0, the convolution matrix $\hat{\mathbf{A}}$ represents the transition probability matrix $\widetilde{\mathbf{A}}\widetilde{\mathbf{D}}^{-1}$ \cite{chiang2019cluster, DBLP:conf/iclr/ZengZSKP20, hamilton2017inductive},
the symmetric normalization adjacency matrix $\widetilde{\mathbf{D}}^{-\frac{1}{2}}\widetilde{\mathbf{A}}\widetilde{\mathbf{D}}^{-\frac{1}{2}}$ \cite{gasteiger_predict_2019, DBLP:conf/iclr/KipfW17} and the reverse transition probability matrix $\widetilde{\mathbf{D}}^{-1}\widetilde{\mathbf{A}}$ \cite{DBLP:conf/icml/XuLTSKJ18}, respectively.

Employing the propagated feature $\mathbf{X}^{(k)}$, the evaluation of node classification task is accomplished by executing the softmax function and optimizing the cross-entropy loss across all labeled nodes.

\subsection{Scalable Graph Neural Networks}

Although GNNs achieve excellent performance by executing multiple feature propagation and transformation processes, it was found that the aggregation of neighbor features (i.e., feature propagation) makes a major contribution to the performance of GNNs and plays a more important role \cite{wu2019simplifying}. Based on this finding, to improve the scalability of GNNs, SGC \cite{wu2019simplifying} was proposed to decompose the two processes and remove feature transformations in the middle layers. It propagates the node features for $k$ times as:

\begin{equation}
\begin{aligned}
&{\mathbf{X}}^{(k)}=\hat{\mathbf{A}}^{k}\mathbf{X},
\end{aligned}
\label{eq_sgc}
\end{equation}
where $\hat{\mathbf{A}}$ is defined as Eq. (\ref{eq_GCN}). 
Then, ${\mathbf{X}}^{(k)}$, the propagated feature at depth $k$, is fed to a linear model for classification. Benefiting from the linear propagation of Eq. (\ref{eq_sgc}), SGC removes the trainable parameters between feature propagation and allows for offline precomputation of the feature matrix as shown in Figure \ref{fig_pre} (b). As a result, SGC effectively reduces training time by avoiding the computationally intensive feature propagation process in each training epoch, since the feature propagation is performed only once before the classifier training.
 
 \begin{figure*}[t]
\centering
\setlength{\abovecaptionskip}{0cm}
\includegraphics[width=0.93\linewidth]{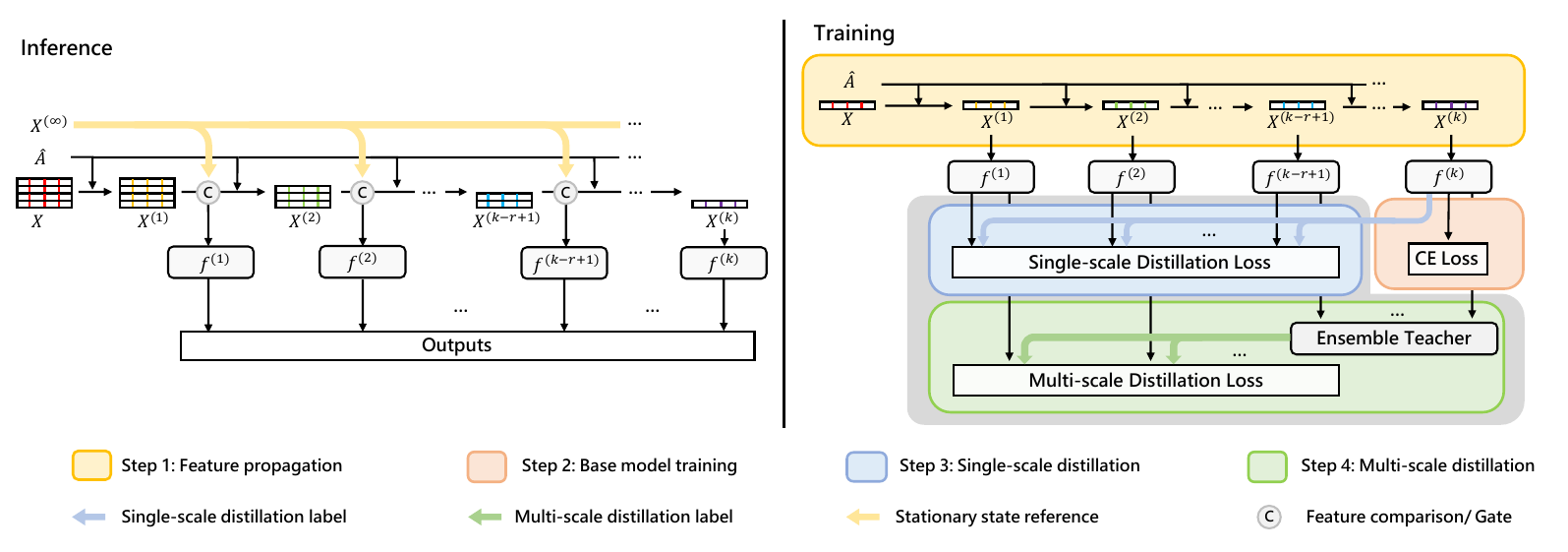}
\caption{The inference procedure and classifier training procedure for NAI. In the inference procedure (left), the propagation depth is adaptively controlled by gates or by comparing the propagated feature with the stationary state. The propagated features with low propagation depth are predicted by corresponding classifier in advance. The training procedure (right) includes feature propagation, base classifier training, and Inception Distillation. The classifier $f^{(k)}$ is initially trained by cross-entropy (CE) loss, and set as the teacher for Single-Scale Distillation. Subsequently, enhanced classifiers at higher depths are employed to construct the ensemble teacher, which serves to distill multi-scale receptive field information into lower depth classifiers.}
\label{fig_main}
\end{figure*}

 Following SGC, more powerful Scalable GNNs are designed by exploring the precomputed features $\{\mathbf{X}^{(0)}, \mathbf{X}^{(1)}, ..., \mathbf{X}^{(k)}\}$ in Eq. (\ref{eq_sgc}), with  $\mathbf{X}^{(0)}$ representing $\mathbf{X}$ for consistent notation. For example, SIGN \cite{sign_icml_grl2020} proposes to transform propagated features at different depths by linear transformations, then concatenates them together to enhance the feature representation. The predicted objective can be represented as: 
 \begin{equation}
\begin{aligned}
&{\mathbf{X}}^{(k)}_{SIGN}={\mathbf{X}}^{(0)}\mathbf{W}^{(0)}\left |  \right |  {\mathbf{X}}^{(1)}\mathbf{W}^{(1)}\left |  \right |  ...\left |  \right |  {\mathbf{X}}^{(k)}\mathbf{W}^{(k)}, 
 \end{aligned}
\label{eq_sign}
\end{equation}
where $ \left |  \right | $ denotes concatenation operations and $\mathbf{W}^{(k)}$ are transformation matrices. $\mathrm{S^2GC}$ \cite{zhu2020simple} averages propagated features at different depths to construct a simple spectral graph convolution: 
\begin{equation}
\begin{aligned}
&{\mathbf{X}}^{(k)}_{S^2GC}=\frac{1}{k} {\textstyle \sum_{l=0}^{k}}{ \mathbf{X}}^{(l)}. 
\end{aligned}
\label{eq_ssgc}
\end{equation}

GAMLP\cite{zhang2022graph} combines propagated features at different depths by measuring the feature information gain and constructing the node-wise attention mechanism:
\begin{equation}
\begin{aligned}
&{\mathbf{X}}^{(k)}_{GAMLP}={\textstyle \sum_{l=0}^{k}} T^{(l)}{\mathbf{X}}^{(l)},
\end{aligned}
\label{eq_gamlp}
\end{equation}
where $T^{(l)}$ are diagonal node-wise attention matrices. 
The propagated features $\{\mathbf{X}^{(0)}, \mathbf{X}^{(1)}, ..., \mathbf{X}^{(k)}\}$ used in these methods are all pre-computed in advance, successfully speeding up the training procedure and transductive inference procedure.
 
However, these Scalable GNNs still suffer from scalability issues when making inferences on the unseen nodes that are not encountered during model training. In such an inductive inference scenario (Figure \ref{fig_pre} (d)), the feature propagation of unseen nodes has to be executed online, and this time-consuming process directly leads to high inference latency.

\section{Methodology}
To mitigate the scalability problem of \textbf{Scalable GNNs} in the \textbf{inductive} inference setting, we propose an online propagation framework, i.e., Node-Adaptive Inference (NAI), as shown in Figure \ref{fig_main}. 
In the inductive inference procedure, the personalized propagation depth is generated online for each test node by referring to its stationary state, and the nodes with low personalized propagation depth are inferred in advance during the propagation procedure.
Within the framework, two optional modules (Distance/Gate-based Node-Adaptive Propagation) are designed  to compare the propagated features with the stationary states and customize the personalized propagation depth.
In the training procedure of multiple classifiers, NAI leverage the Inception Distillation to enhance the classifiers at lower depths. We construct a more powerful teacher to capture multi-scale information of different-sized receptive fields and updates both teacher and students simultaneously with the target of higher prediction accuracy.

In the following sections, we first present the two propagation modules and the inference algorithm in Section \ref{section31}. Subsequently, we discuss the computational complexity of inference in Section \ref{section32}. Finally, we delve into the details of training classifiers for each layer and introduce Inception Distillation in Section \ref{ID}.

\subsection{Node-Adaptive Propagation}
\label{section31}
The key component of the NAI framework is generating the personalized propagation depth in the inference procedure. To this end, we introduce two optional Node-Adaptive Propagation (NAP) modules by referring to the stationary feature state, and the NAP is plugged into the propagation process directly to control the propagation depth for each node. 

Specifically, Scalable GNNs propagate the information within $k$-hops neighbors by multiplying the $k$-th order normalized adjacency matrix by the feature matrix as Eq. (\ref{eq_sgc}). This operation gradually smooths the node feature by neighbor features, and with the growth of $k$, the propagated node features within the same connected component will reach a stationary state \cite{zhang2021node}. When $k\to \infty$, the features are propagated for infinite times and reach the stationary feature state $\mathbf{X^{\left(\infty\right)}}$, which be calculated as:
\begin{equation}
\begin{aligned}
&\mathbf{X}^{\left(\infty\right)}=\hat{\mathbf{A}}^{\left(\infty\right)}\mathbf{X},\\
\end{aligned}    
\label{eq_stationary}
\end{equation}
where $\mathbf{\hat{A}}^{\left(\infty\right)}$ is the adjacency matrix propagated for infinite times. The element in $\mathbf{\hat{A}}^{\left(\infty\right)}$ is calculated as:
\begin{equation}
\begin{aligned}
&\mathbf{\hat{A}}^{\left(\infty\right)}_{i,j}=\frac{\left ( d_i+1 \right ) ^{\gamma}\left (  d_j+1\right ) ^{1-\gamma}}{2m+n},
\end{aligned}    
\label{eq_stationary2}
\end{equation}
where $\mathbf{\hat{A}}^{\left(\infty\right)}_{i,j}$ is the weight between nodes $v_i$ and $v_j$ , i.e., the element of $i$-th row and $j$-th column in $\mathbf{\hat{A}}^{\left(\infty\right)}$. $d_i$ and $d_j$ are node degrees for $v_i$ and $v_j$. $m$ and $n$ are the numbers of edges and nodes. $\gamma$ is the convolution coefficient in Eq. (\ref{eq_GCN}). Since $\mathbf{\hat{A}}^{\left(\infty\right)}_{i,j}$ is only related to the degree of source node $v_i$ and target node $v_j$, topology information is lost after the infinite number of propagation and the final features will be over-smoothed. For example, when $\gamma=0$, $\mathbf{\hat{A}}^{\left(\infty\right)}_{i,j}$ is only determined by the degree of target node $v_j$ and target nodes with equal degrees have the same weight value. 

Considering the different degree distributions of the nodes, the smoothness of the node features varies at the same propagation depth and should be evaluated to avoid the over-smoothing problem. Moreover, the node features with proper smoothness can be inferred in advance to reduce the redundant computation.   
To this end, we propose two different NAP approaches to generate the personalized propagation depth in the explicit and implicit manner respectively: Distance-based NAP ($\rm{NAP_d}$) and Gate-based NAP ($\rm{NAP_g}$).

\subsubsection{\textbf{Distance-based NAP}}
$\rm{NAP_d}$ uses the distance between the propagated feature at depth $l$ (i.e., $\mathbf{X}^{\left(l\right)}_i$) and stationary feature (i.e., $\mathbf{X}^{\left(\infty\right)}_i$) to measure the feature smoothness of node $v_i$ explicitly, and the distance $\Delta^{\left(l\right)}_i$ is defined as Eq. (\ref{eq_distance}):
\begin{equation}
\begin{aligned}
\Delta^{\left(l\right)}_i = \; \left \| {\mathbf{X}}^{\left(l\right)}_i - {\mathbf{X}}^{\left(\infty\right)}_i\right \|,
\end{aligned}    
\label{eq_distance}
\end{equation}
where $\left \| \cdot \right \|$ means $l_2$ norm. A small distance indicates a strong smoothing effect and a higher risk of over-smoothing.  

To control the smoothness and generate a proper propagation depth for inference, we introduce a global hyper-parameter $T_s$ as a threshold for all distances in the graph. The personalized propagation depth $L(v_i, T_s)$ for the node $v_i$ is: 
\begin{equation}
L(v_i, T_s)= \mathop{\arg\min}_{l}( \Delta^{\left(l\right)}_i<T_s).
\label{eq_ppd}
\end{equation}
The personalized propagation depth satisfies the union upper-bound \cite{zhang2021node} as:
\begin{equation}
\footnotesize
\begin{aligned}
L(v_i, T_s)\leq \min\Big\{\log_{\lambda_2}(T_s\sqrt{\frac{d_i+1}{2m+n} } ),   \max\{L(v_j, T_s), v_j\in N_{v_i}\}+1\Big\},
\label{eq_bound}
\end{aligned}  
\end{equation}
where $\lambda_2 \leq 1$ is the second largest eigenvalue of $\hat{\mathbf{A}}$ and $N_{v_i}$ is the neighbor node set of $v_i$.

The first term of upper-bound shows that the personalized propagation depth of $v_i$ is positively correlated with the scale of the graph (i.e., the number of edges $m$ and the number of nodes $n$), the sparsity of the graph (small $\lambda_2$ means strong connection and low sparsity, and vice versa), and negatively correlated with its node degree $d_i$. Moreover, the second term indicates that the difference between two neighboring nodes’ personalized propagation depth is no more than 1, and the neighbors of high-degree nodes will also have a low personalized propagation depth. In summary, nodes located in sparse local areas and possessing smaller degrees should have higher personalized propagation depths, and vice versa.

By referring to the distance explicitly, $\rm{NAP_d}$ takes both the node feature and topological information into account and can generate the personalized propagation depth during the propagation process. With the personalized propagation depth $L(v_i, T_s)$, the propagated feature ${\mathbf{X}}^{\left(L(v_i, T_s)\right)}_i$ will be predicted by classifier $f^{(L(v_i, T_s))}$ (refer to Section \ref{ID} for details) in advance.

\begin{figure*}[ht]
\centering
\setlength{\abovecaptionskip}{0cm}
\includegraphics[width=0.65\linewidth]{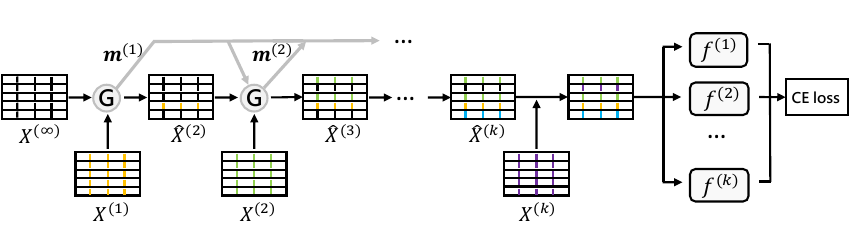}
\caption{The end-to-end training procedure of gates for Scalable GNNs with the highest depth $k$. The propagated features are compared with the stationary feature in each gate and the decisions of gates at higher depths are influenced by previous gates. The propagated features are predicted by corresponding classifiers, and the gates are optimized by using cross-entropy (CE) loss.}
\label{fig_traingate}
\end{figure*}

\subsubsection{\textbf{Gate-based NAP}}
$\rm{NAP_g}$ leverages a series of gates at different depths to control the propagation process and dynamically decide whether the propagation of each node should be stopped. Nonetheless, the gate structure and the training of a series of gates encounter several challenges.

\begin{itemize}
 \item The propagated feature should be compared with the stationary feature in each gate.
 \item Each node should possess a distinct and suitable propagation depth, signifying that it should be chosen by a single gate exclusively.
 \item Gates at varying depths exhibit dependency, as the decisions made by higher-depth gates are influenced by preceding ones. As a result, gates should be trained concurrently to effectively learn this dependency.
 \item Gates must be designed to be lightweight, guaranteeing that no heavy computation operations are incorporated.
\end{itemize}

Considering above challenges, we design gates for Scalable GNNs and simultaneously train these gates end-to-end as show in Figure \ref{fig_traingate} by leveraging the well-trained classifiers (refer to Section \ref{ID} for details). Here we take the propagation depth $l$ as an example.

During the gate training procedure, the gate $g^{(l)}$ at depth $l$ takes the propagated feature $\mathbf{X}^{\left(l\right)}_i$ and $\hat{\mathbf{X}}^{\left(l\right)}_i$ from the previous depth as inputs, subsequently generating a one-hot mask $\boldsymbol{m}^{\left(l\right)}_i$ as output according to Eq. (\ref{gate}). The initial value of $\hat{\mathbf{X}}^{\left(l\right)}_i$ is established as the stationary feature to input the gate $g^{(1)}$, i.e.,  $\hat{\mathbf{X}}^{\left(1\right)}_i=\mathbf{X}^{\left(\infty\right)}_i$.

\begin{equation}
\begin{aligned}
&{\mathbb{X}}^{\left(l\right)}_i= \mathbf{X}^{\left(l\right)}_i \left |  \right |  \hat{\mathbf{X}}^{\left(l\right)}_i, \\
&\mathbf{e}^{\left(l\right)}_i=\mathrm{softmax}\left({{\mathbb{X}}^{\left(l\right)}_i \mathbf{W}^{\left(l\right)}}\right), \\
&\boldsymbol{m}^{\left(l\right)}_i=\mathrm{GS}\left(\mathbf{e}^{\left(l\right)}_i-\mathbf{\Theta}^{\left(l\right)}_i\right),
\label{gate}
\end{aligned}  
\end{equation}
where $1\le l<k$. $ \mathbf{W}^{\left(l\right)}\in \mathbb{R}^{ 2f \times 2}$ is the trainable weight vector. $\mathbf{e}^{\left(l\right)}_i\in \mathbb{R}^{2}$ is the probability which indicates the preference of $\mathbf{X}^{\left(l\right)}_i$ and $\hat{\mathbf{X}}^{\left(l\right)}_i$. $\mathrm{GS}$ is the Gumbel softmax \cite{jang2016categorical} which used to generate two dimensional one-hot mask $\boldsymbol{m}^{\left(l\right)}_i = [{m}^{\left(l\right)}_{i,1}, {m}^{\left(l\right)}_{i,2}]$. 

To prevent a node from being chosen by multiple gates or all nodes employing the propagated feature at the highest depth, we introduce the penalty term $\mathbf{\Theta}^{\left(l\right)}_i$.  This term ensures that each node is selected by the gates only once. Specifically, $\mathbf{\Theta}^{\left(l\right)}_i=[\theta^{\left(l\right)}_{i,1}, 0]$ is generated based on the masks from previous depths and operates exclusively along the first dimension. 
$\theta^{\left(l\right)}_{i,1}= {\textstyle \sum_{j=1}^{l-1}\mu \mathrm{sigmoid}(\varphi({m}^{\left(j\right)}_{i,1}-0.5))}$ in our implementation\footnote{$\mu$ and $\varphi$ are large and arbitrary constants. In our experiments, we set $\varphi$ and $\mu$ equal to 1000.}. For $1\le j \le l-1$, if ${m}^{\left(j\right)}_{i,1}>0.5$ exists, the propagated feature of node $v_i$ has already been selected by previous depth $j$ and a large penalty will be added to $\theta^{\left(l\right)}_{i,1}$. 
Consequently, the mask will maintain ${m}^{\left(l\right)}_{i,1}< {m}^{\left(l\right)}_{i,2}$ for all higher depths $l \le j < k$.
Conversely, if this condition is not met, $\theta^{\left(l\right)}_{i,1}$ will remain at a low value, approximating zero.

With the mask $\boldsymbol{m}^{\left(l\right)}_i$, the gate input at depth $l+1$ is constructed as Eq. (\ref{gate2}):
\begin{equation}
\begin{aligned}
&\hat{\mathbf{X}}^{\left(l+1\right)}_i=\boldsymbol{m}^{\left(l\right)}_i [\mathbf{X}^{\left(l\right)}_i;  \hat{\mathbf{X}}^{\left(l\right)}_i].
\label{gate2}
\end{aligned}  
\end{equation}

\begin{itemize}
 \item If ${m}^{\left(l\right)}_{i,1}> {m}^{\left(l\right)}_{i,2}$, ${\mathbf{X}}^{\left(l\right)}_i$ will be selected by the mask and used as the input of the next depth gate. Owing to the penalty term, the mask for $v_i$ at higher depths will be maintained as ${m}^{\left(l\right)}_{i,1}< {m}^{\left(l\right)}_{i,2}$, which enables the propagated feature at depth $l$ to be predicted in the end.
 \item If ${m}^{\left(l\right)}_{i,1}< {m}^{\left(l\right)}_{i,2}$, $\hat{\mathbf{X}}^{\left(l\right)}_i$ will be chosen by the mask to retain the same value from previous depths. Notice that $\mathbf{X}^{\left(\infty\right)}_i$ is input in the first gate as $\hat{\mathbf{X}}^{\left(1\right)}_i$ and the value of $\hat{\mathbf{X}}^{\left(l\right)}_i$ will be preserved as $\mathbf{X}^{\left(\infty\right)}_i$ if the node has never been selected by any previous gate. 
\end{itemize}

Ultimately, if $\hat{\mathbf{X}}^{\left(k\right)}_i=\mathbf{X}^{\left(\infty\right)}_i$, $\hat{\mathbf{X}}^{\left(k\right)}_i$ will be replaced with $\mathbf{X}^{\left(k\right)}_i$. Nodes selected by different gates will be predicted by the classifiers at their corresponding depths. Employing well-trained classifiers and cross-entropy loss, the gates at various depths are trained concurrently, while classifier parameters remain unaltered during the gate training process.

During the inference procedure, gates will dynamically generate discrete masks according to the propagated feature and stationary feature. The personalized propagation depth for the node $v_i$ is: 
\begin{equation}
\begin{aligned}
&L(v_i)= \mathop{\arg\min}_{l}( {m}^{\left(l\right)}_{i,1}=1).
\label{gateinf2}
\end{aligned}  
\end{equation}

Given the personalized propagation depth $L(v_i)$, the propagated feature ${\mathbf{X}}^{\left(L(v_i)\right)}_i$ will be predicted in advance by classifier $f^{(L(v_i))}$.

\subsubsection{\textbf{Inference Algorithm}}
To unify both $\rm{NAP_d}$ and $\rm{NAP_g}$ into our proposed NAI framework and adapt them to different latency constraints and application scenarios, we introduce two more global hyper-parameters in the inference algorithm, i.e., $T_{min}$ and $T_{max}$, which indicates the minimum and the maximum propagation depth, respectively. The inference procedure is shown in Algorithm \ref{alg:Framwork}. 

In line 2, $\mathbf{X^{\left(\infty\right)}}$ for batch $\mathcal{V}_{b}$ is calculated according the entire graph by Eq. (\ref{eq_stationary}). In line 3, the supporting nodes are derived according to $\mathcal{V}_{b}$ and $T_{max}$, where $1\le T_{max}\le k$. Then, the node features will be propagated $T_{min}$ times, where $1\le T_{min}\le T_{max}$ (line 5). After $T_{min}$ times propagation, features are compared with $\mathbf{X^{\left(\infty\right)}}$ (or input $g^{(l)}$ together with $\mathbf{X^{\left(\infty\right)}}$) and inferred by the classifier if the distances are smaller than $T_{s}$ (or mask is $[1, 0]$) (line 9-12). Until $l=T_{max}$, all left nodes will be classified by $f^{(T_{max})}$ and the prediction results for $\mathcal{V}_{b}$ are output (line 17-18). 

Once the model is deployed on the device, users can choose the hyper-parameters by using validation set that align with the latency requirements and provide the highest validation accuracy for the inference process.

\begin{algorithm}[tb]
  \caption{Inference pipeline of NAI.}
  \LinesNumbered
  \label{alg:Framwork}
   \KwIn{$k$ classifiers, adjacency matrix $\mathbf{\widetilde{A}}$, degree matrix $\mathbf{\widetilde{D} }$, feature matrix $\mathbf{X}$, test set $\mathcal{V}_{test}$, propagation depth $k$, gates, threshold $T_s$, the minimum propagation depth $T_{min}$ and the maximum propagation depth $T_{max}$.}
    \KwOut{The prediction results of $\mathcal{V}_{test}$.}
    \For{\rm{batch} $\mathcal{V}_{b}$ \rm{in} $\mathcal{V}_{test}$}{
    Calculate the stationary feature state $\mathbf{X^{\left(\infty\right)}}$ for $\mathcal{V}_{b}$;\\
    Sample supporting nodes for $\mathcal{V}_{b}$;\\
    \For{$l=1$ \rm{to} $T_{max}$}{
    Calculate the propagated feature ${\mathbf{X}}^{(l)}$ for $\mathcal{V}_{b}$;\\
    \If{$l<T_{min}$}{
        Continue;\\}
    \ElseIf{$l<T_{max}$}{
    \For{$i=1$ \rm{to} $|\mathcal{V}_{b}|$}{
     Calculate the distance $\Delta^{\left(l\right)}_i$ between ${\mathbf{X}_i}^{(l)}$ and $\mathbf{X}_i^{\left(\infty\right)}$ for test node $v_i$ (or generate mask $\boldsymbol{m}^{\left(l\right)}_{i}$ by gate $g^{(l)}$);\\
        \If{$\Delta^{\left(l\right)}_i<T_{s}$ \rm{(or} ${m}^{\left(l\right)}_{i,1}=1$\rm{)}}{
            Predict ${\mathbf{X}_i}^{(l)}$ by classifier $f^{(l)}$;\\
            Remove $v_i$ from $\mathcal{V}_{b}$;\\
            }
        \Else{
    Continue;\\}}}
    \Else{
    Predict $\mathcal{V}_{b}$ by classifier $f^{(l)}$;\\}
    }}
    \Return The prediction results for $\mathcal{V}_{test}$.
\end{algorithm}

\subsection{Computational Complexity Analysis}
\label{section32}
Table \ref{tab_complexity} compares the inference computational complexity of four Scalable GNNs and their complexity after deploying NAI in the inductive setting.
All computations include feature processing and classification, and we show the basic version of GAMLP which utilizes the attention mechanism in feature propagation. 

NAI could reduce the computation of feature propagation by decreasing the propagation depth $k$. Suppose $q$ is the average propagation depth over all nodes when adopting NAI, the complexity for feature propagation in SGC is decreased to $\mathcal{O} (qmf)$. This means that NAI can achieve stronger acceleration effects for graphs with large-scale edges and high feature dimensions under the same $q$. 
The classification complexity is $\mathcal{O} (nf^2)$, which is the same as vanilla SGC. 
The additional computational complexity for the stationary state and distance calculation (or gate) is $\mathcal{O}(n^2f)$ and $\mathcal{O}(qnf)$, respectively.
Similar results can be observed in ${\mathrm{S^2GC}}$ and GAMLP. For SIGN, it concatenates propagated features at different depths before the classification procedure, leading to the increase of feature dimension. 
As a result, the classification computation also decreases from $\mathcal{O}(kPnf^2)$ to $\mathcal{O}(qPnf^2)$ when applying NAI to SIGN. 

\begin{table*}[htb]
\centering
\setlength{\abovecaptionskip}{0cm}
\caption{{The inference computational complexities of Scalable GNNs in the inductive setting. $n$, $m$ and $f$ are the number of nodes, edges, and feature dimensions, respectively. $k$ denotes the propagation depth and $P$ is the number of layers in classifiers. $q$ is the averaged propagation depth when adopting NAI.}}
\label{tab_complexity}
\resizebox{0.8\textwidth}{!}{
\begin{tabular}{l|llll}
\toprule
                   & SGC  & SIGN& ${\mathrm{S^2GC}}$ & GAMLP \\ \hline
Vanilla & $\mathcal{O} (kmf+nf^2)$  &  $\mathcal{O} (kmf+kPnf^2)$  &   $\mathcal{O} (kmf+knf+nf^2)$     &   $\mathcal{O}(kmf+Pnf^2)$     \\
NAI &  $\mathcal{O}(qmf+nf^2+n^2f)$  &  $\mathcal{O}(qmf+qPnf^2+n^2f)$      &   $\mathcal{O}(qmf+qnf+nf^2+n^2f)$  &   $\mathcal{O}(qmf+Pnf^2+n^2f)$ \\ \bottomrule
\end{tabular}}
\end{table*}

\subsection{Enhance Classifiers by Inception Distillation}
\label{ID}
Although the NAP module generates the personalized propagation depth according to topological information and reduces the computational redundancies of feature propagation, the plain classifiers restrict the classification performances. To compensate for the potential inference accuracy loss, we utilize multiple classifiers for the propagated features at different depths and further propose Inception Distillation to exploit the multi-scale receptive field information. The well trained classifiers can directly used in Algorithm \ref{alg:Framwork}. Inception Distillation includes two stages: Single-Scale and Multi-Scale Distillation. 

\subsubsection{\textbf{Single-Scale Distillation}}
For a Scalable GNN with the highest propagation depth $k$, the classifier $f^{(k)}$ is trained firstly with ${\mathbf{X}}^{\left(k\right)}$ by using cross-entropy loss and is designated as the teacher. The outputs of students $f^{(l)}$ ($1\le l<k$) and the teacher $f^{(k)}$ are processed for knowledge distillation as Eq. (\ref{eq_preofflinekdloss}):
\begin{equation}
\begin{aligned}
&\textbf{z}^{(l)}_{i}=f^{(l)}({\mathbf{X}}^{\left(l\right)}_{i}),\\
&\textbf{z}^{(k)}_{i}=f^{(k)}({\mathbf{X}}^{\left(k\right)}_{i}),\\
&\tilde{\boldsymbol{p}}^{(l)}_{i}=\mathrm{softmax}(\textbf{z}^{(l)}_{i}/T),\\
&\tilde{\boldsymbol{p}}^{(k)}_{i}=\mathrm{softmax}(\textbf{z}^{(k)}_{i}/T),\\
\end{aligned} 
\label{eq_preofflinekdloss}
\end{equation}
where $T$ is the temperature, which controls how much to rely on the teacher’s soft predictions \cite{hinton2015distilling}. 
Then, the knowledge of $f^{(k)}$ is distilled to other student classifiers at lower depths $f^{(l)}$ separately as Eq. (\ref{eq_offlinekdloss}). We follow the conventional knowledge distillation \cite{hinton2015distilling} and penalize the cross-entropy loss between the student’s softmax outputs and the teacher’s softmax outputs.
\begin{equation}
\begin{aligned}
&\mathcal{L}^{(l)}_{d}=\frac{1}{\left | \mathcal{V}_{train} \right | } \sum_{v_i\in {\mathcal{V}_{train}}}\ell( \tilde{\boldsymbol{p}}^{(l)}_{i},\tilde{\boldsymbol{p}}^{(k)}_{i}),\\
\end{aligned} 
\label{eq_offlinekdloss}
\end{equation}
where $\ell(\cdot , \cdot)$ is the cross-entropy loss and $\mathcal{V}_{train}$ is the training set. $\left | \cdot \right |$ denotes the set size. 
Besides $\mathcal{L}^{(l)}_{d}$, the node label provides another supervision signal for the students: 
\begin{equation}
\begin{aligned}
&\mathcal{L}^{(l)}_c=\frac{1}{\left | \mathcal{V}_l \right | } \sum_{v_i\in {\mathcal{V}_l}}\ell( \tilde{\boldsymbol{y}}^{(l)}_{i},\boldsymbol{y}_{i}),\\
&\tilde{\boldsymbol{y}}^{(l)}_{i}=\mathrm{softmax}({\textbf{z}}^{\left(l\right)}_{i}),\\
\end{aligned} 
\label{eq_offlineceloss}
\end{equation}
where ${\textbf{z}}^{\left(l\right)}_{i}$ is derived from Eq. (\ref{eq_preofflinekdloss}) and $\boldsymbol{y}_{i}$ is the one-hot label. $\mathcal{V}_{l}$ is the labeled set.

Finally, the Single-Scale Distillation loss $\mathcal{L}^{(l)}_{single}$ is constructed by jointly optimizing $\mathcal{L}^{(l)}_c$ and $\mathcal{L}^{(l)}_{d}$:
\begin{equation}
\begin{aligned}
&\mathcal{L}^{(l)}_{single}=(1-\lambda)\mathcal{L}^{(l)}_c+\lambda T^2\mathcal{L}^{(l)}_{d},\\
\end{aligned} 
\label{eq_offlineloss}
\end{equation}
where $T^2$ is used to adjust the magnitudes of the gradients produced by knowledge distillation \cite{hinton2015distilling} and $\lambda \in [0, 1]$ is the hyper-parameter that balances the importance of two losses.

\subsubsection{\textbf{Multi-Scale Distillation}}
Single-Scale Distillation can enhance the classifiers according to single-scale receptive field information. In order to further enhance the model's representational capacity, an ensemble teacher is constructed to retain multi-scale receptive field information. 
It is voted by $r$ classifiers at higher depths and their predictions $\tilde{\boldsymbol{y}}^{(l)}_{i}$ are combined as: 
\begin{equation}
\begin{aligned}
&\bar{\textbf{z}}_{i}=\mathrm{softmax}(\sum_{l=k-r+1}^{k} {w}^{(l)}_{i}\tilde{\boldsymbol{y}}^{(l)}_{i}),\\
&{w}^{(l)}_{i}=\frac{exp(q^{(l)}_{i})}{\sum_{l=k-r+1}^{k} exp(q^{(l)}_{i})},\\
&{q}^{(l)}_{i}=\delta (\tilde{\boldsymbol{y}}^{(l)}_{i}\textbf{s}^{(l)}),
\end{aligned} 
\label{eq_onlineteacher}
\end{equation}
where $\bar{\textbf{z}}_{i}$ is the ensemble teacher prediction for node $v_i$.
$\delta\left(\cdot\right)$ is the activation function, and we employ the sigmoid function. $\textbf{s}^{(l)}\in \mathbb{R}^{f \times 1}$ is the weight vector which projects the logits into a same subspace to measure self-attention scores. Scalars $q^{(l)}_{i}$ are normalized to weight the predictions $\tilde{\boldsymbol{y}}^{(l)}_{i}$. 

Subsequently, the teacher's knowledge is transferred to student classifiers by optimizing the Multi-Scale Distillation loss $\mathcal{L}^{(l)}_{multi}$:
\begin{equation}
\begin{aligned}
&\mathcal{L}^{(l)}_{multi}=\mathcal{L}_t+(1-\lambda)\mathcal{L}^{(l)}_{c}+\lambda T^2\mathcal{L}^{(l)}_{e},\\
\end{aligned} 
\label{eq_onlineloss}
\end{equation}
where $1 \le l < k$ and $\mathcal{L}^{(l)}_{multi}$ consists of three components: $\mathcal{L}_t$, $\mathcal{L}^{(l)}_{c}$ and $\mathcal{L}^{(l)}_{e}$. $\mathcal{L}_t$ represents the constraint for the ensemble teacher as Eq. (\ref{eq_onlineloss2}):
\begin{equation}
\begin{aligned}
&\mathcal{L}_t=\frac{1}{\left | \mathcal{V}_l \right | } \sum_{v_i\in {\mathcal{V}_l}}\ell(\bar{\textbf{z}}_i,\boldsymbol{y}_i).\\
\end{aligned} 
\label{eq_onlineloss2}
\end{equation}
$\mathcal{L}^{(l)}_{c}$ is the hard label supervision signal defined in Eq. (\ref{eq_offlineceloss}). $\mathcal{L}^{(l)}_{e}$ distills the knowledge from the ensemble teacher to other student classifiers as Eq. (\ref{eq_onlineloss3}): 
\begin{equation}
\begin{aligned}
&\mathcal{L}^{(l)}_{e}=\frac{1}{\left | \mathcal{V}_{train} \right | } \sum_{v_i\in {\mathcal{V}_{train}}}\ell(\tilde{\boldsymbol{p}}^{(l)}_i,\bar{\boldsymbol{p}}_i),\\
&\bar{\boldsymbol{p}}_i=\mathrm{softmax}(\bar{\textbf{z}}_{i}/T),\\
\end{aligned} 
\label{eq_onlineloss3}
\end{equation}
where $\tilde{\boldsymbol{p}}^{(l)}_i$ is derived from Eq. (\ref{eq_preofflinekdloss}).

Notice that not only the student classifiers but also the weight vector $\textbf{s}^{(l)}$ and the ensemble teacher prediction $\bar{\textbf{z}}_{i}$ will be updated simultaneously by optimizing $\mathcal{L}^{(l)}_{multi}$, which provides a trainable regularization term \cite{Yuan_2020_CVPR} with the target of the higher student performance. 
Leveraging the enhanced classifiers obtained through Single-Scale Distillation, Inception Distillation can further capture comprehensive knowledge within multi-scale receptive fields, thereby improving the performance of classifiers at different depths.

\section{Experiments}
We test NAI on real-world graphs with different scales to verify the effectiveness and aim to answer the following six questions. 
\textbf{Q1}: Compared with other state-of-the-art inference acceleration baselines, can NAI achieve better performance? 
\textbf{Q2}: How does each component (e.g., $\rm{NAP_d}$, $\rm{NAP_g}$, and Inception Distillation) in NAI affect the model performance?
\textbf{Q3}: What is the difference between the performance of $\rm{NAP_d}$ and $\rm{NAP_g}$? 
\textbf{Q4}: Can NAI generalize well to different Scalable GNN models? 
\textbf{Q5}: How about the acceleration performance of NAI under different batch sizes? 
\textbf{Q6}: How do hyper-parameters affect NAI?

\subsection{Experimental Settings}

\noindent\textbf{Baselines and Datasets}. We compare our proposed distance-based NAI ($\rm{NAI_d}$) and gate-based NAI ($\rm{NAI_g}$) with the state-of-the-art methods designed for \textbf{GNN inference acceleration}, which include: (1) TinyGNN \cite{yan2020tinygnn}. Distill the knowledge from a deep GNN teacher to a single-layer GNN while exploiting the local structure information within peer nodes. (2) GLNN \cite{zhang2021graph}. Distill the knowledge from a deep GNN teacher to a simple MLP to eliminate the neighbor-fetching latency in GNN inference. Note that GLNN completely abandons the feature propagation to speed up the inference and can be seen as the extremely simplified case of NAI. (3) NOSMOG \cite{tian2022nosmog}. Enhance GLNN by encoding the graph structural information explicitly and utilizing the adversarial feature augmentation to ensure stable learning against noises. (4) Quantization \cite{paszke2019pytorch}. Quantize model parameters from FP32 to INT8.

\begin{table}[ht]
\setlength{\abovecaptionskip}{0cm}
\caption{Datasets properties. $n$, $m$, $f$ and $c$ are the number of nodes, edges, feature dimensions and classes, respectively.}
\label{tab_data}
\resizebox{\linewidth}{!}
{\begin{tabular}{l|rrrrl}
\toprule
Dataset    & $n$ & $m$ & $f$ & $c$ & \#Train/Val/Test\\ \midrule
Flickr     & 89,250   & 899,756   & 500        & 7        & 44k/22k/22k                  \\ 
Ogbn-arxiv    & 169,343   & 1,166,243   & 128        & 40        & 91k/30k/48k                  \\ 
Ogbn-products      & 2,449,029   & 123,718,280  & 100        & 47  &196k/39k/2,213k
\\ \bottomrule
\end{tabular}}
\end{table}

We evaluate our proposed method on three datasets with different scales and characteristics, including: a citation network (Ogbn-arxiv) \cite{DBLP:conf/nips/HuFZDRLCL20}, an image network (Flickr) \cite{DBLP:conf/iclr/ZengZSKP20} and a product co-purchasing network (Ogbn-products) \cite{DBLP:conf/nips/HuFZDRLCL20}. In citation network, papers from different topics are considered as nodes and the edges are citations among the papers. The node attributes are word embedding vectors and each paper's topic is regarded as a node class. Flickr contains descriptions and properties of images and the node class is the image category. In Ogbn-products, the nodes representing products, and edges between two products indicate that the products are purchased together. Node features are generated from the product descriptions and the task is to predict the category of a product. The detailed descriptions of the datasets are provided in Table \ref{tab_data}.\footnote{The dataset split in GLNN \cite{zhang2021graph} and NOSMOG \cite{tian2022nosmog} are different from ours since they further evaluate the method in the transductive setting.}

\noindent\textbf{Evaluation metrics}. The performance of each baseline is evaluated by five criteria, including the accuracy of the test set (ACC), averaged multiplication-and-accumulation operations per node (MACs), averaged feature processing MACs per node (FP MACs), averaged inference time per node (Time) and averaged feature processing time per node (FP time). 

Specifically, MACs for NAI evaluates 4 procedures, including stationary state computation, feature propagation, distance computation (or gate) and classification. Besides these procedures, the Time for NAI further contains the time of supporting node sampling. FP MACs and FP Time for NAI evaluate the feature propagation and distance computation (or gate) procedure.

\begin{table}[tb]
\centering
\setlength{\abovecaptionskip}{0cm}
\caption{{The hyper-parameters for NAI under base model SGC. $*_{single}$ and $*_{multi}$ mean the hyper-parameters for Single-Scale and Multi-Scale Distillation, respectively.}}
\label{tab_para1}
\resizebox{0.8\linewidth}{!}{
\begin{tabular}{l|llll}
\toprule
               & Flickr & Ogbn-arxiv & Ogbn-products \\ \midrule
k                   & 7      & 5          & 5             \\
learning rate     & 0.001  & 0.001      & 0.01          \\
weight decay     & 0      & 0          & 1e-4          \\
dropout           & 0.3    & 0.3        & 0.1           \\
$T_{single}$           & 1.2    & 1          & 1.1           \\
$\lambda_{single}$     & 0.6    & 0.1        & 0.2           \\
$T_{multi}$            & 1.9    & 1.5        & 1             \\
$\lambda_{multi}$         & 0.8    & 0.1        & 0.1          \\  \bottomrule
\end{tabular}}
\end{table}

\begin{table}[tb]
\centering
\setlength{\abovecaptionskip}{0cm}
\caption{{The hyper-parameters for NAI under different models. $*_{single}$ and $*_{multi}$ mean the hyper-parameters for Single-Scale and Multi-Scale Distillation, respectively.}}
 \label{tab_para2}
   \resizebox{0.65\linewidth}{!}{
\begin{tabular}{l|lll}
\toprule
              & $\mathrm{S^2GC}$ & SIGN & GAMLP \\ \midrule
k             & 10                     & 5    & 5     \\
learning rate & 0.001                  & 0.01 & 0.001 \\
weight decay  & 1e-4                   & 1e-5 & 1e-3  \\
dropout       & 0.2                    & 0.1  & 0.1   \\
$T_{single}$         & 1                      & 2    & 1.6   \\
$\lambda_{single}$   & 0.1                    & 0.9  & 0.9   \\
$T_{multi}$          & 1.9                    & 1.8  & 1.8   \\
$\lambda_{multi}$    & 0.6                    & 0.9  & 0.8   \\ \bottomrule
\end{tabular}}
\end{table}

\noindent\textbf{Implementation and Settings}. Without loss of generality, we use the symmetric normalization adjacency matrix $\widetilde{\mathbf{D}}^{-\frac{1}{2}}\widetilde{\mathbf{A}}\widetilde{\mathbf{D}}^{-\frac{1}{2}}$ in all base models. We set all student classifiers to be the same as the teacher GNNs except for GLNN on dataset Ogbn-arxiv and Ogbn-products. We follow GLNN paper and set the hidden embedding size as 4-times and 8-times wider than the teacher GNN on Ogbn-arxiv and Ogbn-products to achieve higher accuracy.  Moreover, for a fair comparison, we re-implement the position feature aggregation in NOSMOG by the matrix multiplication to show NOSMOG’s real inference performance in the inductive setting.\footnote{In the released codes of NOSMOG, the aggregation of position features is calculated node-by-node, which is inefficient and time-consuming in real inference scenarios.}

\begin{table*}[pt]
\setlength{\abovecaptionskip}{0cm}
\caption{{Inference comparison under base model SGC on three datasets. ACC and Time are measured in the percentage and millisecond respectively. Acceleration ratios between NAI and vanilla SGC are shown in brackets.}}
 \label{tab_SGC}
 \resizebox{\textwidth}{!}{
\begin{tabular}{l|lllll|lllll|lllll}
\hline
             & \multicolumn{5}{c|}{Flickr}                               & \multicolumn{5}{c|}{Ogbn-arxiv}                        & \multicolumn{5}{c}{Ogbn-products}                          \\ \hline
             & ACC   & \# mMACs   & \#FP mMACs & Time       & FP Time    & ACC   & \# mMACs  & \#FP mMACs & Time      & FP Time   & ACC   & \# mMACs   & \#FP mMACs & Time        & FP Time    \\ \hline
SGC          & 49.43 & 2475.3     & 2471.2     & 2530.6     & 2381.8     & 69.36 & 895.1     & 887.8      & 1276.7    & 1034.2    & 74.24 & 32946.4    & 32939.7    & 68806.7     & 50628.6    \\
GLNN         & 44.39 & 4.2        & 0.0        & 11.0       & 0.0        & 54.83 & 54.0      & 0.0        & 19.4      & 0.0       & 63.12 & 168.5      & 0.0        & 238.9       & 0.0        \\
NOSMOG       & 48.18 & 4.4        & 0.1        & 28.3       & 17.6       & 67.35 & 7.9       & 0.1        & 31.0      & 29.6      & 72.48 & 7.3        & 0.2        & 73.3        & 60.9       \\
TinyGNN      & 46.80 & 8850.3     & 8846.1     & 1413.8     & 1412.1     & 67.31 & 294.6     & 287.2      & 523.7     & 522.1     & 71.33 & 3418.0     & 3411.3     & 1954.6      & 1948.2     \\
Quantization & 48.34 & 2475.3     & 2471.2     & 2482.2     & 2344.7     & 68.88 & 895.1     & 887.8      & 1223.4    & 1003.6    & 73.01 & 32946.4    & 32939.7    & 68726.0     & 50587.6    \\ \hline
$\rm{NAI_d}$ & 49.36 & 174.9 (14) & 148.3 (17) & 238.5 (11) & 143.4 (17) & 69.25 & 83.5 (11) & 65.1 (14)  & 182.4 (7) & 60.6 (17) & 73.70 & 583.2 (56) & 451.6 (73) & 923.2 (75)  & 591.6 (86) \\
$\rm{NAI_g}$ & 49.41 & 176.1 (14) & 149.4 (17) & 261.9 (10) & 153.7 (15) & 69.34 & 83.8 (11) & 65.4 (14)  & 195.5 (7) & 72.3 (14) & 73.89 & 583.3 (56) & 451.7 (73) & 1088.3 (63) & 602.2 (84) \\ \hline
\end{tabular}}
\end{table*}

\begin{figure*}[ht]
\setlength{\abovecaptionskip}{0cm}
  \centering
  \includegraphics[width=0.85\linewidth]{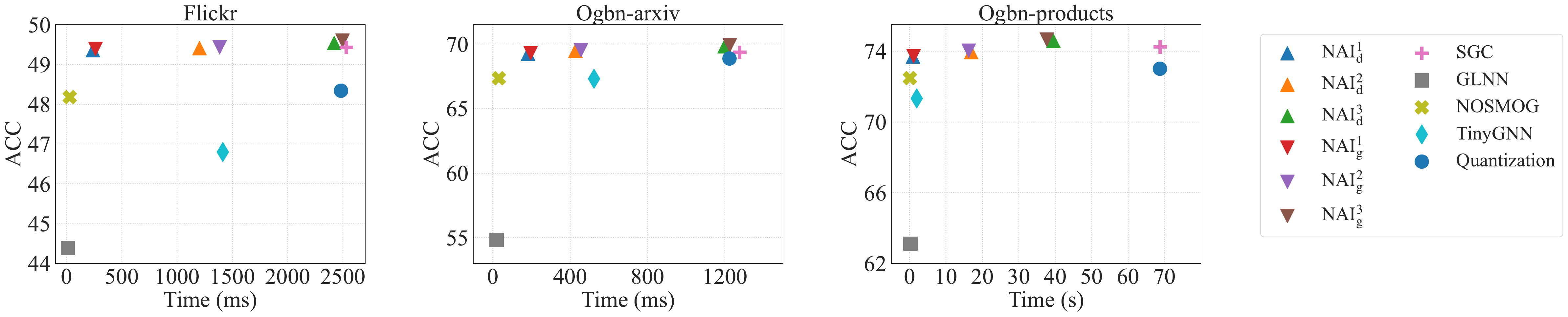}
  \caption{Accuracy and inference time comparison among different baselines and NAIs. NAIs with the superscript indicate 3 different settings.}
  \label{fig_sgc}
\end{figure*}

\begin{table*}[ht]
\centering
\setlength{\abovecaptionskip}{0cm}
  \caption{Node distributions of $\rm{NAI_d}$ and $\rm{NAI_g}$ under three different settings. The propagation depth increases from 1 (left) to $k$ (right).}
  \label{fig_node_distribution}
\resizebox{0.7\textwidth}{!}{
\begin{tabular}{l|lll}
\toprule
               & Fickr                                     & Ogbn-arxiv                         & Ogbn-products                       \\ \midrule
$\rm{NAI^1_d}$ & {[}76,   22237, 0, 0, 0, 0, 0{]}          & {[}1849, 46754, 0, 0, 0{]}         & {[}0, 2213091, 0, 0, 0{]}           \\
$\rm{NAI^2_d}$ & {[}0, 0, 1938, 20375, 0, 0, 0{]}          & {[}0, 20528, 28075, 0, 0{]}        & {[}1086, 0, 2212005, 0, 0{]}        \\
$\rm{NAI^3_d}$ & {[}0,   392, 5580, 848, 85, 308, 15100{]} & {[}0, 16503, 12221, 1077, 18802{]} & {[}0,   1384, 239, 2211468, 0{]}    \\\midrule
$\rm{NAI^1_g}$ & {[}2159, 20154, 0, 0, 0, 0, 0{]}          & {[}2052, 46551, 0, 0, 0{]}          & {[}42514, 2170577, 0, 0, 0{]}       \\
$\rm{NAI^2_g}$ & {[}0,   0, 1373, 12397, 8543, 0, 0{]}     & {[}0, 16420, 32183, 0, 0{]}        & {[}0, 1369671, 843420, 0, 0{]}      \\
$\rm{NAI^3_g}$ & {[}0,   0, 144, 16102, 4477, 1590, 0{]}   & {[}0, 0, 13215, 3864, 31524{]}     & {[}0, 0, 488514, 1301562, 423015{]} \\ \bottomrule
\end{tabular}}
\end{table*}

For each method, the hyper-parameters used in experiments are searched by the grid search method on the validation set or following the original papers. We use the ADAM optimization algorithm to train all the models. The best propagation depth $k$ for each dataset and base model is searched together with learning rate, weight decay, and dropout to get the highest performance. Specifically, the values for $k$, learning rate and weight decay are searched from [2, 10] with step 1, \{0.6, 0.3, 0.1, 0.01, 0.001\} and \{0, 1e-3, 1e-4, 1e-5\}. Dropout, $T$ and $\lambda$ are searching from [0, 0.7], [1, 2] and [0, 1] with step 0.1, respectively. The hyper-parameters of NAI under base model SGC are shown in Table \ref{tab_para1}. Table \ref{tab_para2} shows the hyper-parameters for the NAI under other different base models.

To eliminate randomness, we repeat each method three times and report the mean performance. If not specified otherwise, the inference time is evaluated on the CPU with batch size 500. 
The codes are written in Python 3.9 and the operating system is Ubuntu 16.0. We use Pytorch 1.11.0 on CUDA 11.7 to train models on GPU. All experiments are conducted on a machine with Intel(R) Xeon(R) CPUs (Gold 5120 @ 2.20GHz) and NVIDIA TITAN RTX GPUs with 24GB GPU memory.

\subsection{Performance Comparison}
To answer \textbf{Q1}, we compare $\rm{NAI_d}$ and $\rm{NAI_g}$ with other baselines under the base model: SGC. For NAIs, we select the hyper-parameters that prioritize the inference speed.

From Table \ref{tab_SGC}, we observe that both $\rm{NAI_d}$ and $\rm{NAI_g}$ achieve great balances between accuracy and inference speed. 
As for ACC, NAIs outperform the Quantization method and achieve the least ACC loss compared to vanilla SGC. Although Quantization also shows great accuracy, it only saves the classification computation and could not help to reduce the computation from feature processing. Benefiting from removing the feature propagation in the inference procedure, GLNN has the smallest MACs and the fastest inference speed. However, for the same reason, GLNN could not generalize well for inductive settings as analyzed in their paper. Even with the increased embedding size, the accuracies on Ogbn-arxiv and Ogbn-products decrease significantly. This indicates that ignoring topological information severely impairs the prediction of unseen nodes. Although NOSMOG imitates this problem by using Deepwalk and explicitly encoding the position information, it still has accuracy gaps compared with the base model. Compared to the single-layer GNN, NAIs outperform TinyGNN on all datasets. Although TinyGNN saves a part of the computation of feature propagation, the self-attention mechanism and linear transformation used in its peer-aware module cause a large number of extra computations. Especially in the dataset with high feature dimension, e.g., Flickr, the MACs are much more than vanilla SGC. 

$\rm{NAI_g}$ obtains better accuracy than $\rm{NAI_d}$, which comes at the cost of more computations from gates. Compared with other baselines, $\rm{NAI_d}$ and $\rm{NAI_g}$ accelerate inference significantly by controlling the FP MACs, and $\rm{NAI_d}$ achieves the 75$\times$ Time speedup and 86$\times$ FP Time speedup on Ogbn-products. Although the highest propagation depth is 5 on Ogbn-products,  $\rm{NAI_d}$ and $\rm{NAI_g}$ both achieve nonlinear acceleration ratios since the number of supporting nodes and the adjacency matrix size grow at an exponential rate with the propagation depth. This verifies the effectiveness of generating personalized propagation depth by using the NAI framework.

\begin{table*}[t]
\centering
\setlength{\abovecaptionskip}{0cm}
\caption{{The ablation study on NAPs under different $T_{max}$. The propagation depth of node distribution increases from 1 to $k=5$.}}
 \label{tab_exitnode}
 \resizebox{0.85\linewidth}{!}{
\begin{tabular}{r|l|rrl|rrl}
\toprule
& \multicolumn{1}{c|}{} & \multicolumn{3}{c|}{Ogbn-arxiv} & \multicolumn{3}{c}{Ogbn-products}  \\ \hline
$T_{max}$          & Method                & \multicolumn{1}{r}{ACC (\%)} & Time (ms) & Node distribution                       & \multicolumn{1}{r}{ACC (\%)} & Time (ms) & Node distribution                   \\ \hline
\multirow{3}{*}{2} & NAI w/o NAP           & 69.16 & 202.7 & {[}0, 48603, 0, 0, 0{]}    & 73.70 & 923.2  & {[}0, 2213091, 0, 0, 0{]}      \\
                   & $\rm{NAI_d}$          & 69.25 & 182.4 & {[}1849, 46754, 0, 0, 0{]} & 73.70 & 923.2   & {[}0, 2213091, 0, 0, 0{]}     \\ 
                   & $\rm{NAI_g}$          & 69.34 & 195.5 & {[}2052, 46551, 0, 0, 0{]}  & 73.89 & 1088.3  & {[}42514, 2170577, 0, 0, 0{]} \\
                   \hline
                   
\multirow{3}{*}{3} & NAI w/o NAP           & 69.38 & 454.2  & {[}0, 0, 48603, 0, 0{]}      & 73.95  & 17121.5 & {[}0, 0, 2213091, 0, 0{]}      \\
                   & $\rm{NAI_d}$          & 69.48 & 427.4  & {[}0, 20528, 28075, 0, 0{]}  & 73.97  & 16914.3 & {[}1086, 0, 2212005, 0, 0{]}   \\ 
                   & $\rm{NAI_g}$          & 69.53 & 455.7  & {[}0, 16420, 32183, 0, 0{]}  & 74.03  & 16223.9 & {[}0, 1369671,   843420, 0, 0{]} \\  
                   \hline
\multirow{3}{*}{4} & NAI w/o NAP           & 69.26 & 889.3  & {[}0, 0, 0, 48603, 0{]}          & 74.57  & 42232.2 & {[}0, 0, 0, 2213091, 0{]}      \\
                   & $\rm{NAI_d}$          & 69.52 & 816.6  & {[}0, 30303, 5898, 12402, 0{]}   & 74.58  & 39474.8 & {[}0, 1384, 239, 2211468, 0{]} \\ 
                    & $\rm{NAI_g}$         & 69.66 & 806.4 & {[}0, 31519,   11344, 5740, 0{]}  & 74.66 & 37662.0 & {[}0, 0, 905524, 1307567, 0{]} \\
                   \hline
\multirow{3}{*}{5} & NAI w/o NAP        & 69.36 & 1296.4 & {[}0, 0, 0, 0, 48603{]}            & 74.24  & 68938.8 & {[}0, 0, 0, 0, 2213091{]} \\
                   & $\rm{NAI_d}$       & 69.82 & 1198.9 & {[}0, 16503, 12221, 1077, 18802{]} & 74.58  & 67523.2 & {[}0, 0, 0, 2213068, 23{]}   \\ 
                   & $\rm{NAI_g}$       & 69.90 & 1224.7 & {[}0, 0, 13215,   3864, 31524{]}   & 74.69  & 68087.0 & {[}0, 0, 488514, 1301562, 423015{]}
\\  \bottomrule
\end{tabular}}
\end{table*}

Besides the speed-first results in Table \ref{tab_SGC}, the NAI framework allows users to choose more accurate results based on the latency constraints. 

Figure \ref{fig_sgc} shows the trade-off between accuracy and inference time in different hyper-parameter settings. We select 3 typical settings for each dataset and method, which are denoted as "$\rm{NAI^1_*}$", "$\rm{NAI^2_*}$" and "$\rm{NAI^3_*}$", respectively. Note that "$\rm{NAI^1_*}$" is the speed-first setting in Table \ref{tab_SGC}. From Figure \ref{fig_sgc}, NAIs achieve the highest classification accuracy and are even superior to vanilla SGC. This is due to that NAP mitigates the over-smoothing problem and Inception Distillation enhances the classifiers (Table \ref{tab_exitnode} and \ref{tab_ablation} in the next subsection evaluate their impacts). 
For example, on Flickr, $\rm{NAI^3_d}$ and $\rm{NAI^3_g}$ achieve more accurate results while spending a similar inference time with SGC. $\rm{NAI^2_g}$ speed up $\rm{NAI^3_g}$ by 1.8$\times$, and $\rm{NAI^2_d}$ accelerates $\rm{NAI^3_d}$ by 2.1$\times$ with little accuracy drop. 
More detailed comparisons between $\rm{NAI_g}$ and $\rm{NAI_d}$ can be found in Section \ref{abl_stu}.

Table \ref{fig_node_distribution} shows the detailed test node distributions selected from a single run, i.e., the number of nodes at different propagation depths, for different datasets and hyper-parameter settings. The depth increases from 1 (left) to $k$ (right).
From Table \ref{fig_node_distribution}, we observe that most of the nodes of $\rm{NAI^2_d}$ on Flickr adopt the propagated features at depth 4. This successfully reduces the number of supporting nodes and saves the computation of the feature propagation. To get the best accuracy, $\rm{NAI^3_d}$ makes full use of each classifier, and the propagation depths of tested nodes are various. As for the $\rm{NAI^1_d}$ on Ogbn-products, all nodes adopt the propagated features at depth 2 to trade off the inference speed and accuracy. It demonstrates the flexibility of the $\rm{NAI}$ framework, and the fixed propagation depth used in classic GNNs is the special case of our proposed method.

\subsection{Ablation Study}
\label{abl_stu}

To thoroughly evaluate our method and answer \textbf{Q2-3}, we provide ablation studies on: (1) Node-Adaptive Propagation; (2) Inception Distillation.

Table \ref{tab_exitnode} shows the performance of $\rm{NAI_d}$, $\rm{NAI_g}$ and NAI without NAP under different hyper-parameter settings on Ogbn-arxiv and Ogbn-products. ACC and Time values are averaged over 3 runs and the node distributions are selected from a single run. Their maximum propagation depths $k=5$, and $T_{max}=1$ is omitted due to the same inference accuracy. 
Note that the accuracies of "NAI w/o NAP" do not grow monotonically with $T_{max}$ because the Inception Distillation enhances the classifiers independently. Comparing $\rm{NAI_d}$ with "NAI w/o NAP" under the same $T_{max}$, accuracies are all improved with less inference latency. To achieve a fast inference speed under the same $T_{max}$, tested nodes adopt various propagation depths, contributing to both accuracy improvement and computation saving. $\rm{NAI_g}$ achieves much better accuracy than $\rm{NAI_d}$. This improvement comes from the powerful representation ability of gates and at the cost of the computation and inference time. For example, when $T_{max}=3$, the inference time of $\rm{NAI_g}$ on Ogbn-arxiv is larger than both $\rm{NAI_d}$ and "NAI w/o NAP". However, when more nodes adopt lower propagation depths, $\rm{NAI_g}$ can perform better. When $T_{max}=4$, $\rm{NAI_g}$ is more efficient and effective than $\rm{NAI_d}$.

\begin{table}[pt]
\centering
\setlength{\abovecaptionskip}{0cm}
\caption{{The ablation study on the Inception Distillation. Accuracy (\%) is averaged over 3 runs.}}
 \label{tab_ablation}
  \resizebox{0.8\linewidth}{!}{
\begin{tabular}{l|rrrrrr}
\toprule
    & Flickr      & Ogbn-arxiv  & Ogbn-products \\ \hline
NAI w/o ID   & 40.86  &65.54    &70.17   \\
NAI w/o MS   & 44.41  &65.91    &70.28  \\
NAI w/o SS  & 42.81  &66.08    &70.37  \\
NAI          & 44.85    &66.10      &70.49 \\ \bottomrule
\end{tabular}}
\end{table}

Besides NAP, Inception Distillation is designed to explore multi-scale knowledge and improve the inference accuracy. We evaluate the accuracy of $f^{(1)}$, which has the worst performance among classifiers, to show the effectiveness of each component in Inception Distillation. Table \ref{tab_ablation} displays the results of NAI without Inception Distillation ("w/o ID"), NAI without Single-Scale Distillation ("w/o SS"), NAI without Multi-Scale Distillation ("w/o MS") and NAI. First, the Multi-Scale Distillation explores multi-scale receptive fields and constructs a more powerful teacher via self-attention mechanism, contributing to improvements on all datasets when comparing NAI with NAI w/o MS. For example, when ignoring the Multi-Scale Distillation, the accuracy of NAI will drop 0.44\% on Flickr. Besides, Single-Scale Distillation provides a solid foundation for Multi-Scale Distillation. With more accurate classifiers, the ensemble teacher will be more expressive and powerful, which could provide higher-quality supervision signals. The classification results will decrease on all datasets when Single-Scale Distillation is removed. With the help of Single-Scale Distillation, the accuracy of Multi-Scale Distillation has a 2.04\% increase on the dataset Flickr. These results indicate that Single-Scale and Multi-Scale Distillation are essential to NAI.

\subsection{Generalization}

In addition to SGC, our proposed NAI framework can be applied to any Scalable GNNs. To answer \textbf{Q3}, we test the generalization ability of $\rm{NAI_d}$ and $\rm{NAI_g}$ by deploying them on SIGN, $\mathrm{S^2GC}$ and GAMLP. The hyper-parameters, including the classifier structure, are searched to get the best performance for each base model. The detailed hyper-parameters are listed in Table \ref{tab_para2}.

\begin{table}[tp]
\setlength{\abovecaptionskip}{0cm}
\caption{{Inference comparison under base model SIGN on Flickr. ACC is evaluated in percentage. Time and FP Time are evaluated in millisecond. Acceleration ratios between NAI and vanilla GNNs are shown in brackets.}}
\label{tab_basemodel1}
\resizebox{\linewidth}{!}{
\begin{tabular}{l|rrrrr}
\toprule
             & ACC      & \#mMACs & \#FP mMACs & Time    & FP Time    \\ \hline
SIGN     &  51.00 & 1574.9 & 1526.8    & 1667.1  & 1569.0        \\
GLNN            &  46.84 & 8.1   & 0       & 7.8      & 0      \\
NOSMOG          & 48.24 & 8.5 & 0.1 & 32.8 & 17.2           \\
TinyGNN         &   47.21 & 8862.2 & 8846.1    & 1356.1 & 1345.9      \\ 
Quantization    & 45.87 & 1574.9 & 1526.8    & 1654.3  & 1565.0       \\\hline
$\rm{NAI_d}$    &  51.02 & 135.0 (12)  & 112.5 (14)     & 170.4 (10)   & 78.7 (20)   \\
$\rm{NAI_g}$    & 50.93 & 135.2 (12) & 112.8 (14) & 181.3 (9) & 86.6 (18)    \\ \bottomrule
\end{tabular}}
\end{table}

\begin{table}[t]
\setlength{\abovecaptionskip}{0cm}
\caption{{Inference comparison under base model $\mathrm{S^2GC}$ on Flickr. ACC is evaluated in percentage. Time and FP Time are evaluated in millisecond. Acceleration ratios between NAI and vanilla GNNs are shown in brackets.}}
\label{tab_basemodel2}
\resizebox{\linewidth}{!}{
\begin{tabular}{l|rrrrr}
\toprule
             & ACC      & \#mMACs & \#FP mMACs & Time    & FP Time     \\ \hline
$\mathrm{S^2GC}$      & 50.08 & 3897.8 & 3889.2    & 3959.5 & 3717.6               \\
GLNN             & 46.59 & 8.6    & 0       & 9.5     & 0                \\
NOSMOG           & 48.19 & 9.0 & 0.1 & 31.6 & 17.3                   \\
TinyGNN          & 46.89 & 8855.1 & 8846.5    & 1366.7 & 1355.0             \\ 
Quantization     & 49.10 & 3897.8 & 3889.2    & 3946.9 & 3714.6             \\\hline
$\rm{NAI_d}$     & 48.94 & 120.1 (32) & 89.0 (44)   & 149.9 (26)   & 86.3 (43)  \\
$\rm{NAI_g}$     & 49.66 & 142.1 (27) & 97.7 (40) & 165.6 (24) & 88.0 (42)      \\ \bottomrule
\end{tabular}}
\end{table}

\begin{table}[t]
\setlength{\abovecaptionskip}{0cm}
\caption{{Inference comparison under base model GAMLP on Flickr. ACC is evaluated in percentage. Time and FP Time are evaluated in millisecond. Acceleration ratios between NAI and vanilla GNNs are shown in brackets.}}
\label{tab_basemodel3}
\resizebox{\linewidth}{!}{
\begin{tabular}{l|rrrrr}
\toprule
             & ACC      & \#mMACs & \#FP mMACs & Time    & FP Time    \\ \hline
GAMLP        & 51.18 & 1594.8 & 1590.6    & 1759.6  & 1657.6             \\
GLNN            & 46.99 & 8.6   & 0          & 9.2    & 0              \\
NOSMOG          & 48.41 &  9.0  &  0.1       &  31.5  &  16.9            \\
TinyGNN         & 47.40 & 8875.8 & 8873.7    & 1389.1 & 1381.8         \\ 
Quantization    & 50.81 & 1594.8 & 1590.6    & 1701.6  & 1650.8        \\\hline
$\rm{NAI_d}$    & 50.89 & 150.0 (11)  & 124.9 (13)     & 220.1 (8)   & 133.3 (12)   \\
$\rm{NAI_g}$    & 51.04 & 158.4 (10) & 130.9 (12) & 235.2 (7) & 140.3 (12)     \\ \bottomrule
\end{tabular}}
\end{table}
The accuracy and inference time results of SIGN, $\mathrm{S^2GC}$ and GAMLP are shown in Table \ref{tab_basemodel1}, \ref{tab_basemodel2} and \ref{tab_basemodel3}, respectively. NAIs consistently outperform the other baselines when considering both accuracy and inference speedup. Compared to GLNN, $\rm{NAI_d}$ can improve the accuracy for 4.18\%, 2.35\% and 3.90\% on SIGN, $\mathrm{S^2GC}$, and GAMLP, respectively. Although the attention mechanism used in TinyGNN requires a large number of MACs, the feature propagation is more time-consuming and the acceleration ratios for different base models are ranging from 1.2$\times$ to 2.9$\times$ compared with vanilla GNNs. Quantization achieves the smallest accuracy loss but the acceleration ratio is limited. When applying $\rm{NAI_d}$ to SIGN, $\mathrm{S^2GC}$ and GAMLP, the FP Time can be accelerated by $20\times$, $43\times$ and $12\times$. Considering the other computations, i.e., the computation of stationary state and classification, the corresponding inference time are accelerated by $10\times$, $26\times$ and $8\times$. The best acceleration result of $\rm{NAI_g}$ is shown in $\mathrm{S^2GC}$. It achieves $24\times$ and $27\times$ speedup for inference time and MACs, respectively.

\subsection{Effect of Batch Size}

The inference of GNNs is related to the batch size because the number of supporting nodes grows with the increase of the batch size. To answer \textbf{Q4}, we evaluate different methods and report the averaged MACs and inference time in different batch sizes on Flickr. The batch size changes from 100 to 2000, and all other settings and accuracy performances are the same as Table \ref{tab_SGC}. 

\begin{figure}[t]
\centering
\setlength{\abovecaptionskip}{0cm}
\includegraphics[width=0.9\linewidth]{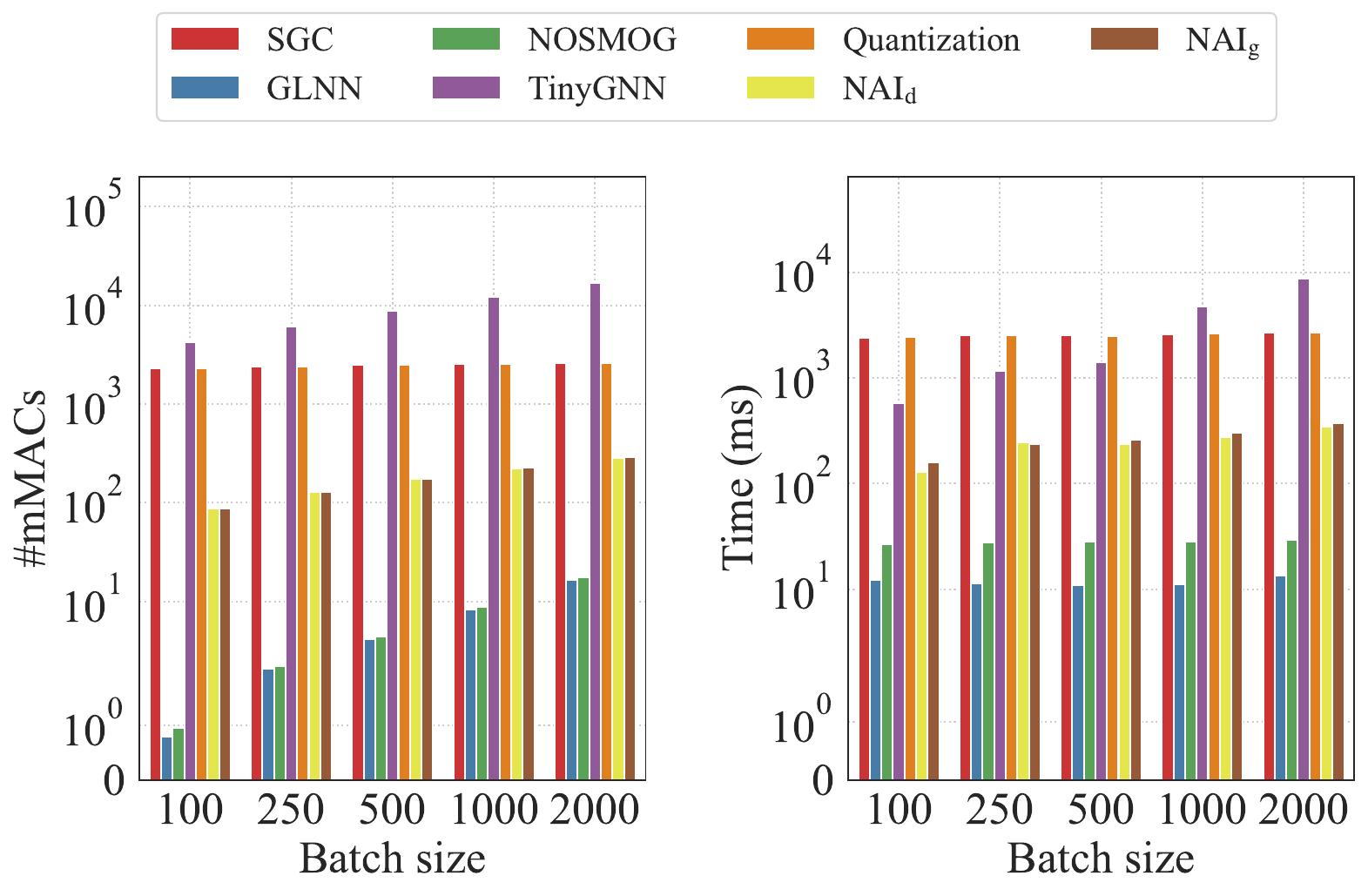}
\caption{The number of MACs and inference time comparison among different methods and batch sizes on the dataset Flickr.}
\label{fig_batch}
\end{figure}

The detailed results are shown in Figure \ref{fig_batch}. For the base model SGC, both MACs and inference time keep at the same magnitude with the growth of batch size. Specifically, when the batch size changes from 100 to 2000, the MACs increase from 2282.8 to 2613.0,  and the inference time increases from 2396.0 ms to 2701.1 ms. The Quantization has the same performance trend as the base model in both MACs and inference time. However, TinyGNN shows a strong positive correlation with the batch size and the inference time exceeds SGC when the batch size is 1000. When the batch size grows to 2000, the number of MACs reaches 16917.9 and the inference time is 8725.0 ms. This performance degradation of TinyGNN comes from the heavy computations of the attention mechanism used in its peer-aware module. Benefiting from the extremely simplified MLP model, the number of MACs of GLNN is kept around 100, and the inference time is around 10 ms. For NOSMOG, the position feature aggregation for unseen nodes causes the MACs and inference time overhead compared with GLNN. For our proposed $\rm{NAI_d}$ and $\rm{NAI_g}$, the number of MACs increases with the growth of batch size. This is due to that NAI requires the extra calculation of the stationary feature state $\mathbf{X^{\left(\infty\right)}}$ and distances (or gates) for all target nodes in the batch. However, these procedures can be fast calculated by matrix multiplication and subtraction, resulting in stable inference time performances in different batch sizes.

\subsection{Parameter Sensitivity Analysis}
Temperature $T$ and weight $\lambda$ are two influential hyper-parameters for Inception Distillation. Moreover, the ensemble number $r$ controls the teacher quality in Multi-Scale Distillation. To analyze the influence of these hyper-parameters and answer \textbf{Q5}, we conduct the experiment on Flickr and the base model is SGC.
\begin{figure}[t]
\centering
\setlength{\abovecaptionskip}{0cm}
\includegraphics[width=0.9\linewidth]{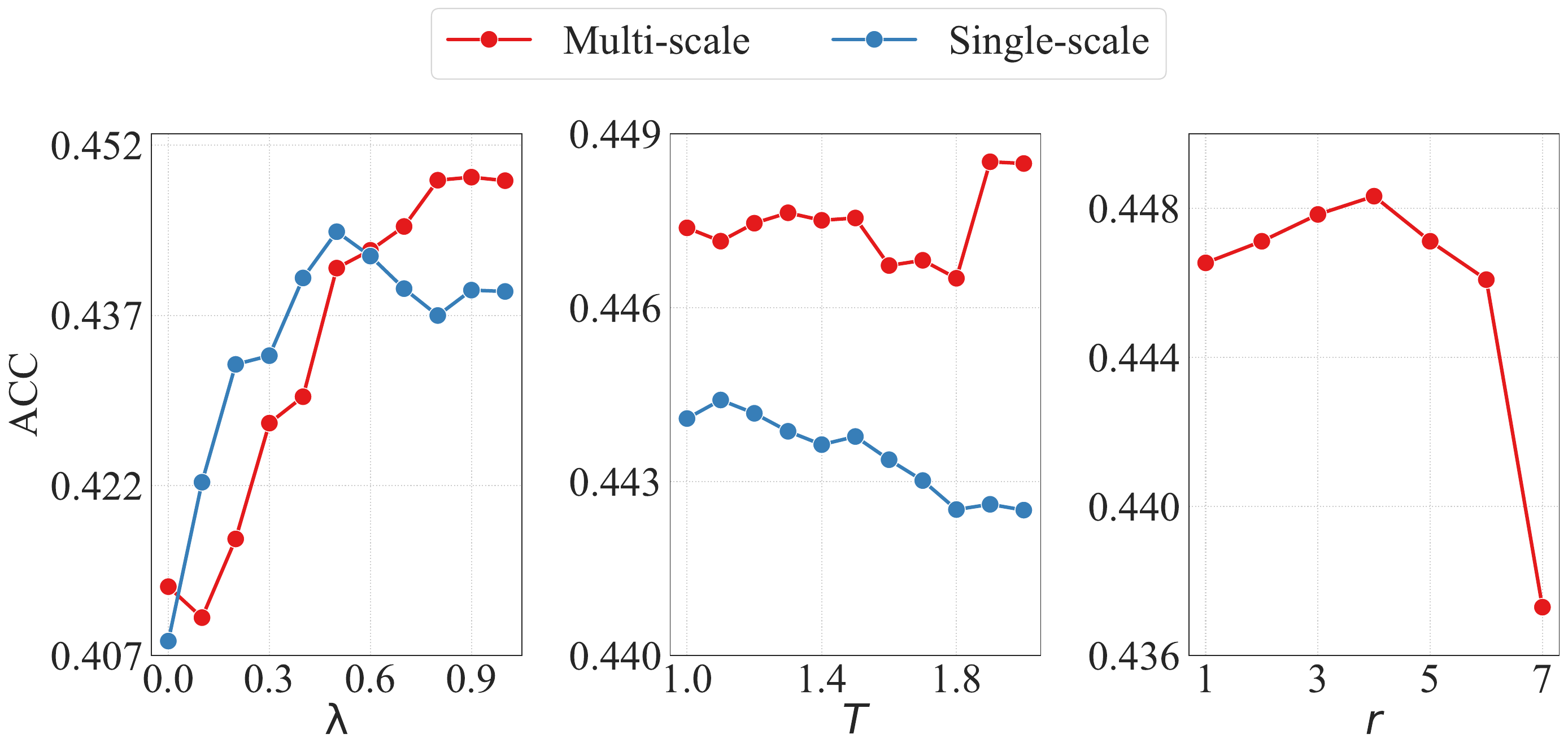}
\caption{Parameter sensitivity results of NAIs under the base model SGC on Flickr.}
\label{fig_para}
\end{figure}
The classification performances of $f^{(1)}$ in terms of hyper-parameters are shown in Figure \ref{fig_para}. Firstly, $\lambda$ is quite important which could significantly affect the classification result. For example, $\lambda$ for Multi-Scale Distillation should be controlled between 0.8 and 1 to get better performance. This indicates that the supervision provided by the ensemble teacher is more important than the hard label. In contrast, $\lambda$ for Single-Scale Distillation should be selected carefully to balance two losses. 
Following the increase of $T$, the performance of Multi-Scale Distillation decreases first and then increases. Thus, limiting $T$ to a larger value and using softer labels works best. The Single-Scale Distillation results in terms of $T$ show that decreasing temperature could help enhance the classification performance. $T$ should be controlled in the range of [1, 1.2]. 
Finally, the results in terms of $r$ show that increasing the number of combined predictions could help enhance the classification performance. But it also introduces more unreliable labels in model training. Especially when introducing the low-quality labels from $f^{(1)}$, the classification result drops rapidly. 

To sum up, Inception Distillation gets stable and high classification performances when $\lambda$ ranges from 0.5 to 1. Softer labels and an appropriate ensemble number should be applied to Multi-Scale Distillation for better performance.

\section{Related Works}
To deploy the model on large-scale graphs, researchers propose various techniques to accelerate training and inference, which can be categorized into the model perspective and the algorithm perspective. 

\noindent\textbf{Acceleration models}. From the model perspective, the acceleration model mainly contain sampling-based and scalable models. Besides the models studied in this paper, sampling-base models can be divided into three categories according to sampling methods: node-wise \cite{hamilton2017inductive, chen2017stochastic, bojchevski2020pprgo}/ layer-wise \cite{DBLP:conf/iclr/ChenMX18, huang2018adaptive, zou2019layer}/ graph-wise \cite{chiang2019cluster, DBLP:conf/iclr/ZengZSKP20} sampling. Although sampling-based GNNs mitigate the neighbor explosion problem by restricting the number of neighbors, they are greatly influenced by sampling quality and suffer from the high variance problem when applied to inference. As a result, the inference procedure typically adheres to conventional GNNs. A distinct sampling method for GNNs acceleration is PPRGo \cite{bojchevski2020pprgo}. PPRGo utilizes the personalized PageRank to replace the hierarchical feature propagation and then proposes an approximate method to select top-k neighbor nodes and speed up the PageRank matrix calculation. Compared with our NAI, PPRGo focuses on the different GNN framework which conducts the propagation process after the transformation process \cite{gasteiger2018predict} to solve the over-smoothing problem. This makes PPRGo must be trained end-to-end and could not be generalized to Scalable GNNs studied in this paper. Instead of using neighbor sampling or matrix approximate calculation, NAI can adaptively generate personalized propagation depth for each node and achieves great generalization ability. 

\noindent\textbf{Acceleration algorithms}. From the algorithm perspective, acceleration methods include pruning, quantization and knowledge distillation (KD). Pruning methods designed for GNNs \cite{zhou2021accelerating, chen2021unified, hui2023rethinking} reduce the dimension of embeddings in each hidden layer to save the computation. Quantization \cite{tailor2020degree} uses low-precision integer arithmetic during inference to speed up the computation. However, these two kinds of methods concentrate on reducing the computation of feature transformation and classification, and raw features are preserved to avoid performance degradation. This limits the acceleration performance considering that feature propagation accounts for the most proportion of runtime. KD aims to train a lightweight model which has a similar performance to the teacher model, thus has been exploited in recent inference acceleration researches. Most KD methods for GNNs try to enhance the student performance by introducing high-order structural information because the receptive field is bound to the number of GNNs layers \cite{yang2020distilling, jing2021amalgamating, yang2021extract, yanggeometric}. Besides, GraphAKD \cite{he2022compressing} leverages adversarial training to decrease the discrepancy between teacher and student. ROD \cite{zhang2021rod} uses multiple receptive field information to provide richer supervision signals for sparsely labeled graphs. RDD \cite{zhang2020reliable} defines the node and edge reliability to make better use of high-quality data. Different from the above works which concentrate on improving the performance of a single classifier, the Inception Distillation in NAI focuses on multi-scale knowledge transfer and boosts the performance for multiple students.

\section{Conclusion}
We present Node-Adaptive Inference (NAI), a general inference acceleration framework for Scalable GNNs. NAI can successfully reduce the redundancy computation in feature propagation and achieve adaptive node inference with personalized propagation depths. With the help of Inception Distillation, NAI exploits multi-scale receptive field knowledge and compensates for the potential inference accuracy loss. Extensive experiments on large-scale graph datasets verified that NAI has high acceleration performance, good generalization ability, and flexibility for different latency constraints. NAI drives the industrial applications of Scalable GNNs, especially in streaming and real-time inference scenarios.
\section*{Acknowledgment}
This work is supported by the Australian Research Council under the streams of Future Fellowship (No. FT210100624) and Discovery Project (DP240101108), as well as 
the high-performance computing platform of Peking University.
\bibliographystyle{IEEEtran}
\bibliography{IEEEabrv,ref}

\end{document}